\documentclass{article}

\usepackage{graphicx}
\usepackage{amsmath}
\usepackage{amsthm}
\usepackage{amssymb}
\usepackage{xcolor}
\usepackage[natbibapa,nodoi]{apacite}
\usepackage{url} 
\usepackage{hyperref}

\usepackage{etoolbox}
\newtoggle{publish}
\toggletrue{publish}

\begin{document}

\newtheorem{property}{Property}[section]
\newtheorem{theorem}{Theorem}[section]
\newtheorem{lemma}[theorem]{Lemma}
\newtheorem{proposition}[theorem]{Proposition}
\newtheorem{corollary}[theorem]{Corollary}

\newcommand{\domain}{\cal D}
\newcommand{\DegreeSecondMoment}{\left< k^2 \right>}

\title{Ease of dependency distance minimization in star-like structures}
\author{Emília Garcia-Casademont and Ramon Ferrer-i-Cancho}

\maketitle

\iftoggle{publish}{}
{
\input{conventions}
\input{pending}
}

\begin{abstract}
The syntactic structure of a sentence can be represented as a tree where edges indicate syntactic dependencies between words. When that structure is a star, 
it has been demonstrated that the head should be placed in the middle of the linear arrangement  according to the principle of syntactic dependency distance minimization. However, hubs of stars tend to be put at one of the ends, against that principle. Here we address two questions: (1) How difficult is it to minimize dependency distance? (2) Why anti dependency distance minimization effects have been found in star structures but not in path structures?  
The ease of optimization is determined by the shape of the optimization landscape. It was demonstrated that the landscape of star structures is quasiconvex (Ferrer-i-Cancho 2015, Language Dynamics and Change). As for (1), here we show that it is indeed convex (a particular case of quasiconvexity) both for star trees and quasistar trees and thus the distance-based optimization problem is simpler than previously believed. As for (2), we argue that (a) competing principles, rather than the difficulty of optimization, must be the actual reason for anti-dependency distance minimization effects and that (b) dependency distance minimization on star-like structures is less rewarding compared to other structures.
\end{abstract}

\iftoggle{publish}{}
{
\tableofcontents
}

\section{Introduction}

\begin{figure}
\centering
\includegraphics[width = \textwidth]{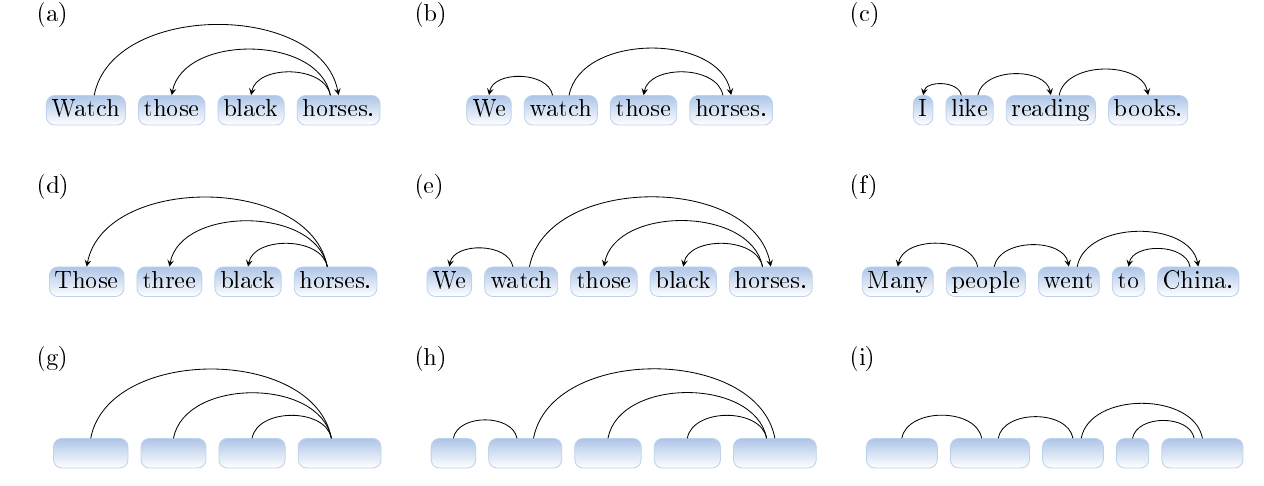}
\caption{\label{fig:sentences} Dependency structures with $n = 4$ or $n=5$. 
(a). A star dependency structure.
(b). A path or quasistar dependency structure.
(c). A path or quasistar dependency structure.
(d). A star dependency structure.
(e). A quasistar dependency structure.
(f). A path dependency structure.
(g). The common underlying free tree for the dependency structures in (a) and (d).
(h). The underlying free tree for the dependency structure in (e).
(i). The underlying free tree for the dependency structure in (f).
}
\end{figure}

The syntactic dependency structure of a phrase or a sentence can be represented as a tree where edges indicate syntactic dependencies between words (Figure \ref{fig:sentences} (a)-(f)). Syntactic dependency structures are rooted trees, namely trees such that a single node, the root, has no incoming dependencies, and all edges are oriented away from the root. 
Hereafter we use $n$ to refer to the number of vertices of a syntactic dependency structure ($n = 4$ in figure \ref{fig:sentences} (a)-(d); $n=5$ in figure \ref{fig:sentences} (e)-(f)).

A rooted tree has an underlying free tree, that results from neglecting link direction. Figure \ref{fig:sentences} (g) shows the underlying free tree of Figure \ref{fig:sentences} (a) and (d).
Figure \ref{fig:sentences} (h) shows the underlying free tree of Figure \ref{fig:sentences} (e).
Figure \ref{fig:sentences} (i) shows the underlying free tree of Figure \ref{fig:sentences} (f).
The free tree of a syntactic dependency structure can be a star tree as in Figure  \ref{fig:sentences} (g). 
A star tree can be defined by means of the concept of vertex degree, i.e. the number of connections of a vertex in the free tree. A star tree is formed by a hub vertex, the vertex that has degree $n -1$ (that is ``horses'' in Figure \ref{fig:sentences} (a) and (d)) and $n - 1$ leaf vertices, namely vertices that have degree one. 
A star tree corresponds to two kinds of syntactic dependency structures, one in which the root is the hub, producing a single-head structure, e.g., a noun and its dependents (Figure \ref{fig:sentences} (d)), and another in which the root is a leaf, producing a structure where there are two heads: the root (that is a leaf) and the hub (Figure \ref{fig:sentences} (a)).

In general, trees vary concerning their degree of star-likeness, namely how much they resemble a star tree with respect to concentration of edges on a few vertices. An approach is hubiness, the variance of the degree distribution, that can be measured by the second moment of degree about zero, i.e. \citep{Ferrer2013b}
\begin{equation}
\DegreeSecondMoment = \frac{1}{n} \sum_{i=1}^n k_i^2
\end{equation}
where $k_i$ is the degree of the $i$-th vertex.
Given $n$, the largest value of $\DegreeSecondMoment$ (i.e. $n - 1$) is achieved by a star tree, the 2nd largest value is achieved by a quasistar tree while the minimum value is achieved by a path tree \citep{Ferrer2016d}. 
A quasistar of $n$ vertices is formed by connecting a new vertex to one of the leaves of a star of $n - 1$ vertices. Figure \ref{fig:sentences} (b), (c) and (e) show quasistar structures. 
A quasistar has a hub of degree $n - 2$ and $n - 2$ leaves.
A path is a tree such that the maximum degree is 2. A path tree has two leaves and the remainder of the vertices have degree 2. 
Figure \ref{fig:sentences} (b), (c) and (f) show path structures.

Figure \ref{fig:sentences} (a) shows a dependency structure such that the hub is the root whereas Figure \ref{fig:sentences} (d) shows one where the hub is not. When the tree structure is a star, the hub vertex is likely to be the root across languages \citep{Ferrer2024d}. When the tree structure is a quasistar, the hub vertex is less likely to be the root but still above chance level.

\begin{figure}
\centering
\includegraphics[width = 0.4\textwidth]{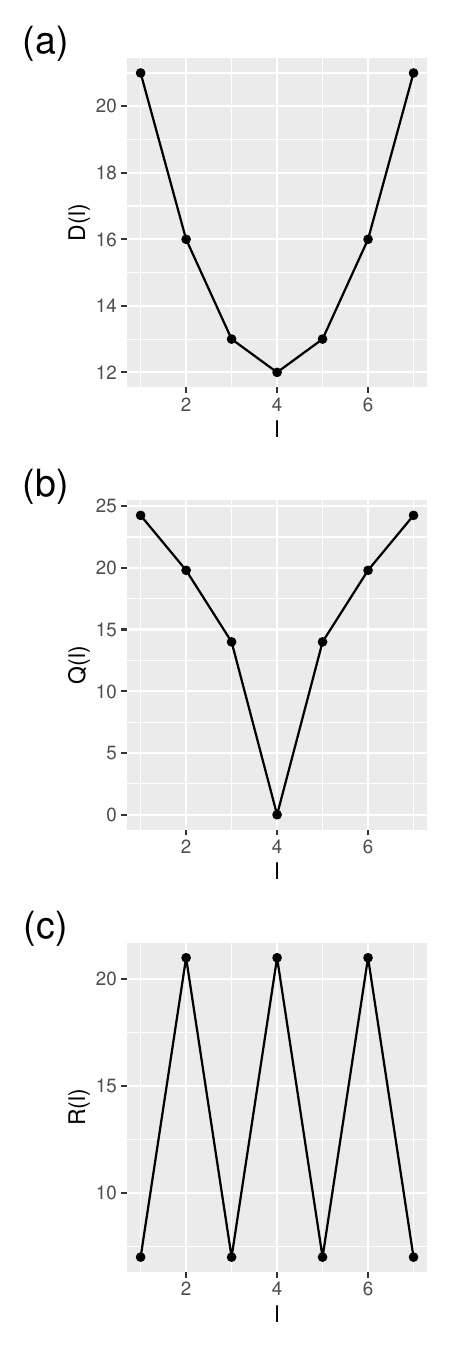}
\caption{\label{fig:convexity} Discrete functions on $l$ within the interval $[1, n]$ with $n = 7$. (a). A convex function that is defined by a polynomial of degree 2 (Equation \ref{eq:total_distance_identity}). (b). A quasiconvex function that is not convex and that is defined as $Q(l) = 2n(l^* - l)^{1/2}$ with $l^* = \lceil n/2 \rceil$. (c). A function defined as $R(l) = n(2 + (-1)^{l})$. }
\end{figure}

The distance between two vertices in positions $i$ and $j$ in the linear arrangement is the absolute value of the difference between the positions, i.e. $|i-j|$. Thus consecutive vertices are at distance 1. Vertices separated by exactly one vertex in between are at distance 2 and so on. Consider a linear arrangement of the vertices of a star tree. 
Consider $D(l)$, the total sum of dependency distances of a star tree when the hub is placed in position $l$. Figure \ref{fig:sentences} (a) and (d) show linear arrangements of star trees with $l = 4$ and $D(l) = 1 + 2 + 3 = 6$.
$D(l)$ is a polynomial of degree 2 of $l$, that is \citep{Ferrer2013e}
\begin{equation}
D(l) = l^2 + (n+1)\left[\frac{1}{2}n - l \right].
\label{eq:total_distance_identity}
\end{equation}
as illustrated in Figure \ref{fig:convexity} (a).

There is a long tradition that regards the total sum of dependency distances of a syntactic dependency structure as a measure of cognitive cost \citep{Liu2017a,Temperley2018a}. However, the implicit assumption is that the cognitive cost of processing a dependency is proportional to distance, which may not be true. To overcome this limitation, it was assumed that the cognitive cost of a link between two vertices at distance $d$ is $g(d)$, a strictly monotonically increasing function of $d$ \citep{Ferrer2013e}. In a star structure, the total cost, $D(l)$, can be decomposed into $D_-(l)$, i.e. the contribution to $D(l)$ of dependencies of edges formed with leaves that precede the hub, and $D_+(l)$, i.e. the contribution of dependencies of edges formed with leaves that follow the hub, as follows 
\begin{eqnarray*}
D_-(l) & = & \sum_{d=1}^{l - 1} g(d) \\
D_+(l) & = & \sum_{d=1}^{n - l} g(d).
\end{eqnarray*}
Then $D(l)$ can be expressed as \citep{Ferrer2013e} 
\begin{eqnarray}
D(l) & = & D_-(l) + D_+(l) \nonumber \\
     & = & \sum_{d=1}^{l - 1} g(d) + \sum_{d=1}^{n - l} g(d).
\label{eq:total_cost}
\end{eqnarray}
When $g$ is the identity function, i.e. $g(d) = d$, one retrieves equation \ref{eq:total_distance_identity}.

Interestingly, the optimal placement of the hub, namely the position of the hub in the linear arrangement that minimizes $D(l)$ is independent of $g$ provided that $g$ is strictly monotonically increasing. If $n$ is odd, the hub must be placed in the only middle position; if $n$ is even, the hub must be placed in one of the two middle positions \citep{Ferrer2013e}. \footnote{See Theorem 2.1 (Online memory cost of the dependencies) by \citet{Ferrer2013e} for further mathematical detail. }
Furthermore, $D(l)$ increases as the hub is moved from the optimal position towards one of the ends. Thus, the placement of the head in Figure \ref{fig:sentences} (a) and (d) is anti-optimal, i.e. it maximizes $D(l)$. 

In spite of the wide evidence of dependency distance minimization in languages \citep{Liu2008a,Futrell2015a,Ferrer2020b},
evidence in distinct contexts suggests a bias to place the hub of star dependency structures at one of the ends against the minimization of the syntactic dependency distance 
\begin{enumerate}
\item
\citet{Courtin2019a} investigated dependency structures in subsequences with $n = 3$ and classified them into four types: chain (the hub is put in the middle and is the root), balanced chain (the hub is put in the middle but is not the root), bouquet (the hub is put at one of the ends and is the root) and zigzag (the hub is put at one of the ends but is not the root). The types that put the hub at one of the ends had a frequency greater than expected by chance: bouquet structures in French, Chinese and English and zigzag structures in Japanese.
\item
\citet{Ferrer2019a} found languages where dependency distances were longer than expected in sequences with $n = 3$ and $n = 4$. When $n = 4$, the trees can only be paths or stars. Interestingly, the effect was only detected in stars.
\item 
\citet{Ferrer2023b} confirmed the finding of anti dependency distance minimization in short sequences by means of a novel optimality score.
\item
\citet{Ferrer2023b} investigated the preferred order in a nominal phrase with star structure across languages. The phrase is formed by a nominal head and three dependents: adjective, numeral and demonstrative. Languages tend to prefer a placement of the nominal hub at one of the ends.
\end{enumerate} 

The robust evidence of anti dependency distance minimization
raises at least three questions: 
\begin{enumerate}
\item[a)]
Which other optimization forces may surpass or counteract the principle of syntactic dependency distance minimization? It has been demonstrated mathematically that the principle of dependency distance minimization is in conflict with the principle of predictability maximization (or surprisal minimization) \citep{Ferrer2013f,Ferrer2024a}. In star structures, the principle of dependency distance minimization predicts that the hub should be placed in the middle of the linear arrangement while the principle of predictability maximization predicts that the hub should be placed at one of the ends.
\item[b)]
How difficult is it to minimize dependency distance costs in star-like structures? So far, ease of optimization has been investigated in stars \citep{Ferrer2013e}. What about quasistars, the structures with the second highest hubiness after stars?
\item[c)]
Why anti dependency distance minimization effects have been found in star structures but not in path structures \citep{Ferrer2019a}? 
\end{enumerate}  

The goal of the present article is to shed light on Questions b) and c). 

The ease of optimization depends on the shape of the word order optimization landscape \citep{Wright1932a,Wright1967}. For simplicity, we assume that only dependency distances matter. 
In previous research, it has been demonstrated that such a unidimensional landscape is quasiconvex \citep{Ferrer2013e}. In particular, it has been shown that $D(l)$ is a quasiconvex function, namely if we have three distinct possible placements of the head, $l_1$, $l_2$, and $l_3$, such that $1\leq l_1 < l_2 < l_3 \leq n$, 
\begin{equation*}
D(l_2) \leq \max(D(l_1), D(l_3)). 
\end{equation*}
Figure \ref{fig:convexity} (a) and (b) show landscapes that are quasiconvex. Figure \ref{fig:convexity} (c) shows a landscape that is not quasiconvex.
If $D(l)$ were a ``convex'' function, that would imply that the minimization of $D(l)$ would be easier compared to quasiconvexity \citep{Boyd2009a}.

The remainder of the article is organized as follows.
Section \ref{sec:convexity} addresses Question b) and shows that both stars and quasistars have a 
``convex'' dependency distance landscape. 
Section \ref{sec:quadratic_cost} addresses Question c) by showing that dependency distance minimization on star-like structures is not rewarding enough compared to other structures. While the cost $D$ is always quadratic on $n$ for star and quasistar structures, independently of the linear arrangement, other structures, e.g. paths, can achieve a subquadratic cost when dependency distance is minimized.  
Section \ref{sec:discussion} reviews the findings and concludes that
competing principles and the inherent quadratic cost of star-like structure, rather the difficulty of optimization, must be the actual reason for anti-dependency distance minimization in star structures.

\section{Convexity}
\label{sec:convexity}

First, note that the popular definition of convexity is typically applied to continuous functions \citep{Boyd2009a}.
For simplicity, suppose a function of a single variable as $D(l)$, that is $f: \domain \rightarrow \mathbb{R}$, where $\domain \subseteq \mathbb{R}$ is the domain of the function. $f$ is convex if and only if for all $\alpha \in [0,1]$ and all $x_1$, $x_2 \in \domain$, 
\begin{equation}
f(\alpha x_1 + (1-\alpha)x_2) \leq \alpha f(x_1) + (1 - \alpha) f(x_2).
\label{eq:convex_function}
\end{equation}
As $D(l)$ is a function of a single variable that is defined on a discrete domain, the standard notion of ``convexity'' needs to be adapted.
In section \ref{sec:star_trees}, we show that $D(l)$ is a ``convex'' function under different notions of convexity for discrete variables. In section \ref{sec:quasistar_trees}, we extend the arguments to quasistar structures and show that quasistar trees also define a convex landscape.

\subsection{Star trees}
\label{sec:star_trees}

We consider approaches t
o the ``convexity'' of $D(l)$ of increasing complexity and strength.

\subsubsection{Convex sequence}

One approach is to regard $D(l)$ as the sequence
\begin{equation*}
D(1), D(2),...,D(n)
\end{equation*}
and then apply the notion of convex sequence \citep{ElFarissi2024a}. \footnote{What follows is borrowed almost verbatim from \citet{ElFarissi2024a}.}
A sequence $(a_i)_{i \geq 1}$ of real numbers (also noted as $(a_i)_{i \in \mathbb{N}}$), i.e.  
\begin{equation*}
a_1, a_2, ..., a_i,...
\end{equation*}
is said to be convex if and only if it satisfies that, for every $i \geq 2$,
\begin{equation}
a_i \leq \frac{a_{i-1} + a_{i+1}}{2}.
\label{eq:convex_sequence}
\end{equation}
The sequence in Figure \ref{fig:convexity} (a) is convex while the sequence in Figure \ref{fig:convexity} (b) is quasiconvex but not convex.
The definition in Equation \ref{eq:convex_sequence} can be derived by setting $x_1$, $x_2$ and $\alpha$ in Equation \ref{eq:convex_function} to $i - 1$, $i + 1$ and $1/2$ and thus one may think that Equation \ref{eq:convex_sequence} leads to a less restrictive version of ``convexity'' for discrete functions. 

Put differently, why just evaluating a local property, a property on each three consecutive points (equation \ref{eq:convex_sequence}) would suffice to conclude a global property, namely that the whole sequence is convex? The key is that
there is a bidirectional relationship between a convex function of one variable and a convex sequence: if $f$ is a convex function defined on $[1, \infty)$, then ${f(i)}_{i \geq 1}$ is a convex sequence. Conversely, if a sequence $(a_i)_{i \geq 1}$ is convex then the function $f$ whose graph is the polygonal line with corner points 
$(i, a_i)$ is a convex function on $[1, \infty)$ (see Remark 1.12 by \citet{Pecaric1992a} and \citet{Adamovic1989a}). Figure \ref{fig:convexity} shows examples of these polygonal lines: the black circles are the points and the segments that join consecutive points define the polygonal line. \\

A natural definition of global convexity applied to discrete sequences derives from the one-to-one mapping of discrete sequences to polygonal lines. A discrete sequence is globally convex if for any two non-consecutive indices $i, k$ with $i < k$, the segments of the sequence between them lie below or on the line connecting the endpoints $(i, x_i)$ and $(k, x_k)$ in the corresponding polygonal line, sometimes referred to as the \textit{Secant Line definition}.

Hence, for discrete sequences it holds that local convexity implies global convexity. \\

The following property indicates that $D(l)$ satisfies equation \ref{eq:convex_sequence} and hence is a convex sequence according to the previous definition.

\begin{property}
$D(l)$ (Equation \ref{eq:total_cost}), where $g(d)$ is a strictly monotonically increasing function of $d$, defines a convex sequence for $l \in [1, n]$.
\label{prop1}
\end{property}

\begin{proof} 
To prove convexity, we must show that the local condition (equation \ref{eq:convex_sequence})
\begin{equation*}   
D(l) \leq \frac{D(l-1) + D(l+1)}{2},
\end{equation*} 
or its equivalent $D(l-1) + D(l+1) \geq 2D(l)$, holds. 

We apply the general definition of $D(l)$ in equation \ref{eq:total_cost} to express $D(l-1)$ and $D(l+1)$ in terms of $D(l)$ as
\begin{align*}
D(l-1) &= \sum_{d=1}^{l-2} g(d) + \sum_{d=1}^{n - l + 1} g(d) \\ \
&= \left( \sum_{d=1}^{l-1} g(d) - g(l-1) \right) + \left( \sum_{d=1}^{n - l} g(d) + g(n - l + 1) \right) \\ \
&= D(l) - g(l-1) + g(n - l + 1) \\
D(l+1) &= \sum_{d=1}^{l} g(d) + \sum_{d=1}^{n - l - 1} g(d) \\
&= \left( \sum_{d=1}^{l-1} g(d) + g(l) \right) + \left( \sum_{d=1}^{n - l} g(d) - g(n - l) \right) \\
&= D(l) + g(l) - g(n - l)
\end{align*}
Now, summing the two expressions:
\begin{align*}
D(l - 1) + D(l + 1) &= 2D(l) + \underbrace{\left[ g(l) - g(l - 1) \right]}_{\geq 0} + \underbrace{\left[ g(n - l + 1) - g(n - l) \right]}_{\geq 0}
\end{align*}
Since $g(d)$ is a strictly monotonically increasing function of $d$, the differences in brackets are both positive (i.e. $\geq 0$) for all valid $l$, hence $D(l - 1) + D(l + 1) \geq  2D(l)$, proving the strict convexity of the sequence.
\end{proof} 

\subsubsection{Discrete convexity}

Beyond the unidimensional scenario, a more general approach to convexity for a discrete function is due to \citet{Yuceer2002a}, who proposes a definition for the $n$-dimensional case using a condition defined on any pair of points as in equation \ref{eq:convex_function}.

\citet{Yuceer2002a} also defines convexity by extending the concept of convexity from continuous functions to discrete functions, in this case, to the general case of functions defined in a subspace of an $n$-dimensional space. In this more general framework, the definition of discrete convexity presented using equation \ref{eq:convex_sequence}
 emerges as a specific case.

Discrete convexity is defined as follows. A function $f: S \rightarrow \mathbb{R} $ defined on $S$, a subspace of a discrete $n$-dimensional space, is discretely convex if for all pairs of points $x$ and $y$ in $S$ and for all $\alpha \in (0,1)$ 
\begin{equation*}
\alpha f(x) + (1-\alpha) f(y) \geq \min\limits_{u \in N(z)} f(u),     
\end{equation*}
where 
\begin{eqnarray*}
N(z) & = & \{u \in S |~||u-z|| < 1\} \\
z & =& \alpha x + (1-\alpha) y \\ 
||u|| & = & \max\limits_{1\leq i \leq n}\{|u_i|\}.    
\end{eqnarray*} \\
In the case of a univariate function, the \textit{Secant Line Definition} is equivalent to the above definition of discrete convexity.

Seemingly, the discrete convexity of a univariate function can also be proved using the non-decreasing condition of the first forward differences \citep[Theorem 1]{Yuceer2002a}
\begin{equation}
\Delta f(x+1) \geq \Delta f(x),
\label{eq:theorem1}
\end{equation}
where $\Delta f(x) = f(x+1) - f(x)$.

\begin{property}
\label{prop:non-decreasing_first_forward_differences}
Consider the definition of $D(l)$ in equation \ref{eq:total_cost}) with $g(d)$ being a strictly monotonically increasing function of $d$. $D(l)$ has non-decreasing first forward differences within $l \in [1, n]$, i.e. 
\begin{equation*}
\Delta D(l+1) \geq \Delta D(l)
\end{equation*}
for $l \in [1, n - 2]$.
\end{property}

\begin{proof}
We have
\begin{align*}
   \Delta D(l+1) &= D(l+2) - D(l+1) \\ 
   &= \left(\sum\limits_{d=1}^{l+1} g(d) + \sum\limits_{d=1}^{n-l-1}g(d)\right)  - \left(\sum\limits_{d=1}^{l}g(d) + \sum\limits_{d=1}^{n-l}g(d)\right) \\
   &= g(l+1) - g(n-l)  \\
    \Delta D(l) &= D(l+1) - D(l) \\
        &= \left(\sum\limits_{d=1}^{l}g(d) + \sum\limits_{d=1}^{n-l}g(d)\right) - \left(\sum\limits_{d=1}^{l-1}g(d) + \sum\limits_{d=1}^{n+1-l}g(d)\right) \\
        &= g(l) - g(n+1-l) \\ 
        \end{align*}
and then  $\Delta D(l+1) \geq \Delta D(l)$ because      
         $$D(l + 1) = g(l+1) - g(n-l) \geq D(l) = g(l) - g(n+1-l)$$ 
and $g$ is strictly monotonically increasing. 
\end{proof}

In conclusion, the function $D(l)$ (equation \ref{eq:total_cost}) is discretely convex because it satisfies equation \ref{eq:convex_sequence} and also equivalently equation \ref{eq:theorem1}.

\paragraph{Strong discrete convexity}

The concept of Strong Discrete Convexity in the literature often aligns with properties found in Discrete Convex Analysis (DCA), such as $\mathrm{L}^\natural$-convexity \citep{chen2017lnatural, murota2001lconvex}, which guarantees even stronger algorithmic properties than simple discrete convexity. For a discrete function $f: S \to \mathbb{R}$ on an $n$-dimensional space, a stringent definition is provided by the following conditions \citep{Yuceer2002a}:

\begin{enumerate}
\item Local Submodularity, i.e.
\begin{equation}
f(x + u) + f(x) \geq f(x \vee u) + f(x \wedge u)   \label{eq:loca_submodularity} 
\end{equation} 
where $u = (u_1,.., u_n) \neq 0, u_i = 0,-1,+1$
for each $i=1,2,..., n$ and $x \vee u=(\max\{x_i, x_i +u_i\})$, $x \wedge u=(\min\{x_i; x_i +u_i\})$. This condition is crucial for proving tractability in multi-dimensional optimization.
\item Monotonicity of Aggregate First Differences, i.e.

$$\sum\limits_{j=1}^n \Delta_j f(x) \leq \sum\limits_{j=1}^n \Delta_j f(x + e_i)$$ 
for all $i = 1, 2, \dots, n$,

where the first forward difference of $f$ at $x$ is

\[
\Delta_i f(x) = f(x + e_i) - f(x),
\]

 or an equivalent condition on non-negativity of the aggregate second differences:

\[
\sum\limits_{j=1}^n \Delta_{ij} f(x) \geq 0
\]

where the second forward differences are defined as:

\[ \Delta_{ij} f(x) = \Delta_i(\Delta_j f(x)) = f(x + e_i + e_j) - f(x + e_j) - f(x + e_i) + f(x). \]

for any coordinates $i$ and $j$

\end{enumerate}

In the one-dimensional case (as for $D(l)$), both conditions simplify significantly. 

\begin{property}
\label{strong-convex}
$D(l)$ (equation \ref{eq:total_cost}), with $g(d)$ being a strictly monotonically increasing function of $d$, defines a strongly discrete function  for $l \in [1, n]$.
\end{property}

\begin{proof}
The second condition (Monotonicity of Aggregate First Differences) is equivalent to the convexity we have already proven (Property \ref{prop:non-decreasing_first_forward_differences}), and equivalent to equation \ref{eq:theorem1}. The first condition (Local Submodularity) simplifies to checking the  condition for $u=\pm 1$, which trivially leads to an equality in equation \ref{eq:loca_submodularity}. Therefore, the function $D(l)$ is indeed strongly discretely convex.
\end{proof}

\subsection{Quasistar trees} 
\label{sec:quasistar_trees}

A quasistar tree of $n$ vertices is formed by connecting a new vertex to one of the leaves of a star of $n - 1$ vertices. This way one obtains one hub of degree $n-1$, one vertex of degree 2, and 
$n - 2$ leaves.
See images (c) and (e) in figure \ref{fig:sentences}.

\begin{figure}
\includegraphics[width = \linewidth]{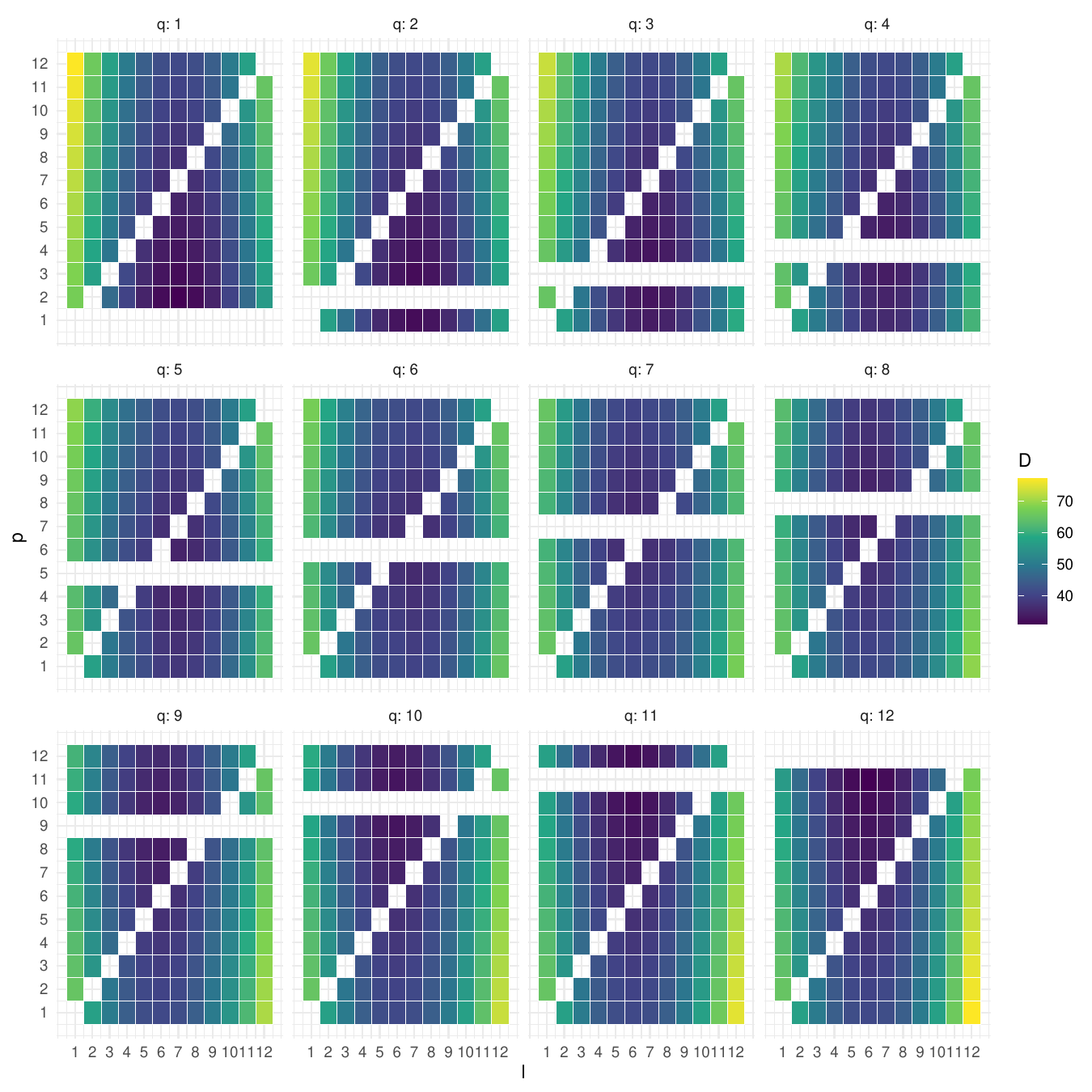}
\caption{\label{fig:landscape_quasistar_by_q} The total cost of dependency distances ($D$) of a quasistar tree with $n = 12$ as a function of $l$, $p$ and $q$, assuming $g(d) = d$. Each panel corresponds to a slice of the 4D heatmap that is defined by the value of $q$. Holes represent impossible combinations of $l$, $p$ and $q$. }
\end{figure}

\begin{figure}
\includegraphics[width = \linewidth]{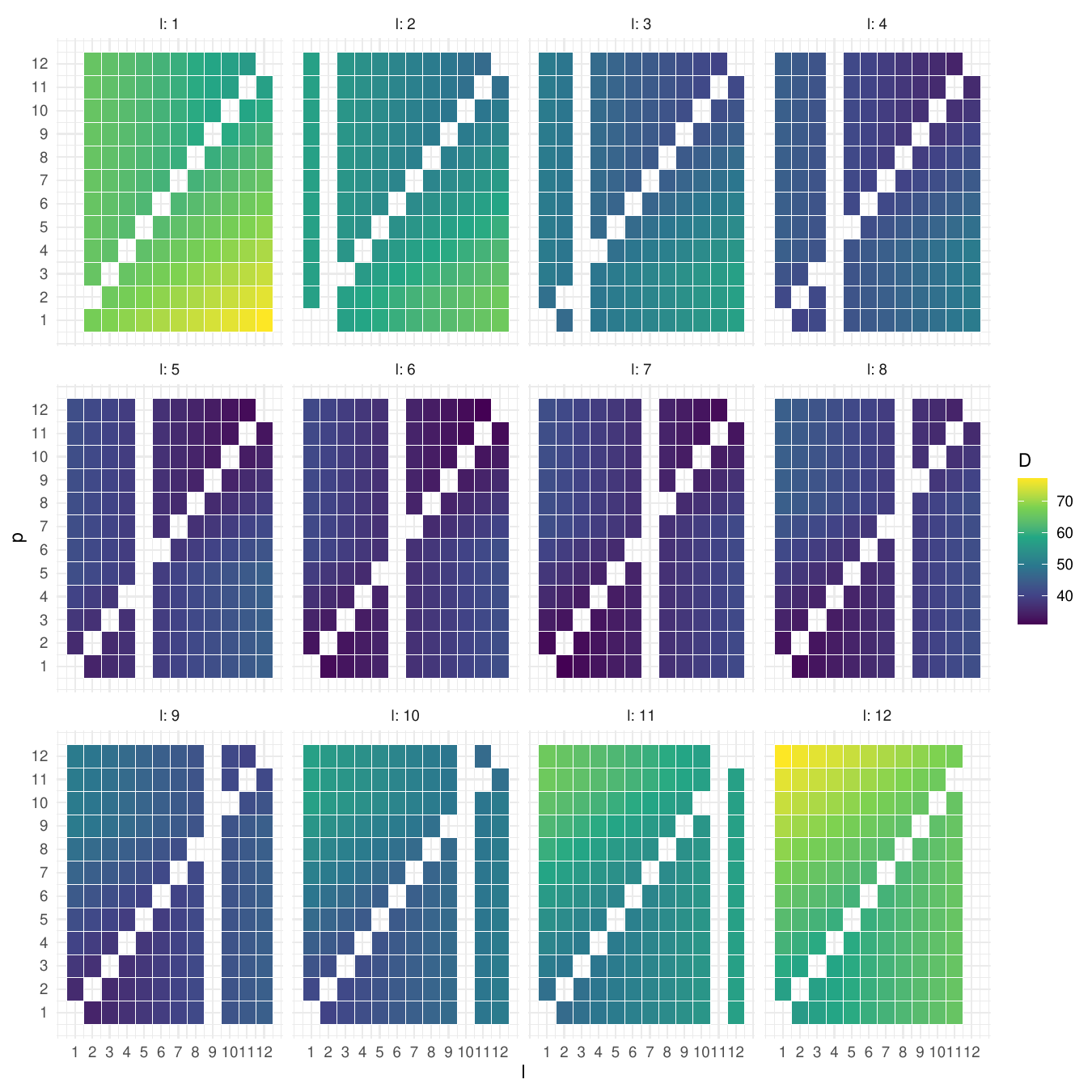}
\caption{\label{fig:landscape_quasistar_by_l} The total sum of dependency distances ($D$) of a quasistar tree with $n = 12$ as a function of $l$, $p$ and $q$. Each panel corresponds to a slice defined by the value of $l$. The format is the same as in Figure \ref{fig:landscape_quasistar_by_q}. }
\end{figure}

\begin{figure}
\includegraphics[width = \linewidth]{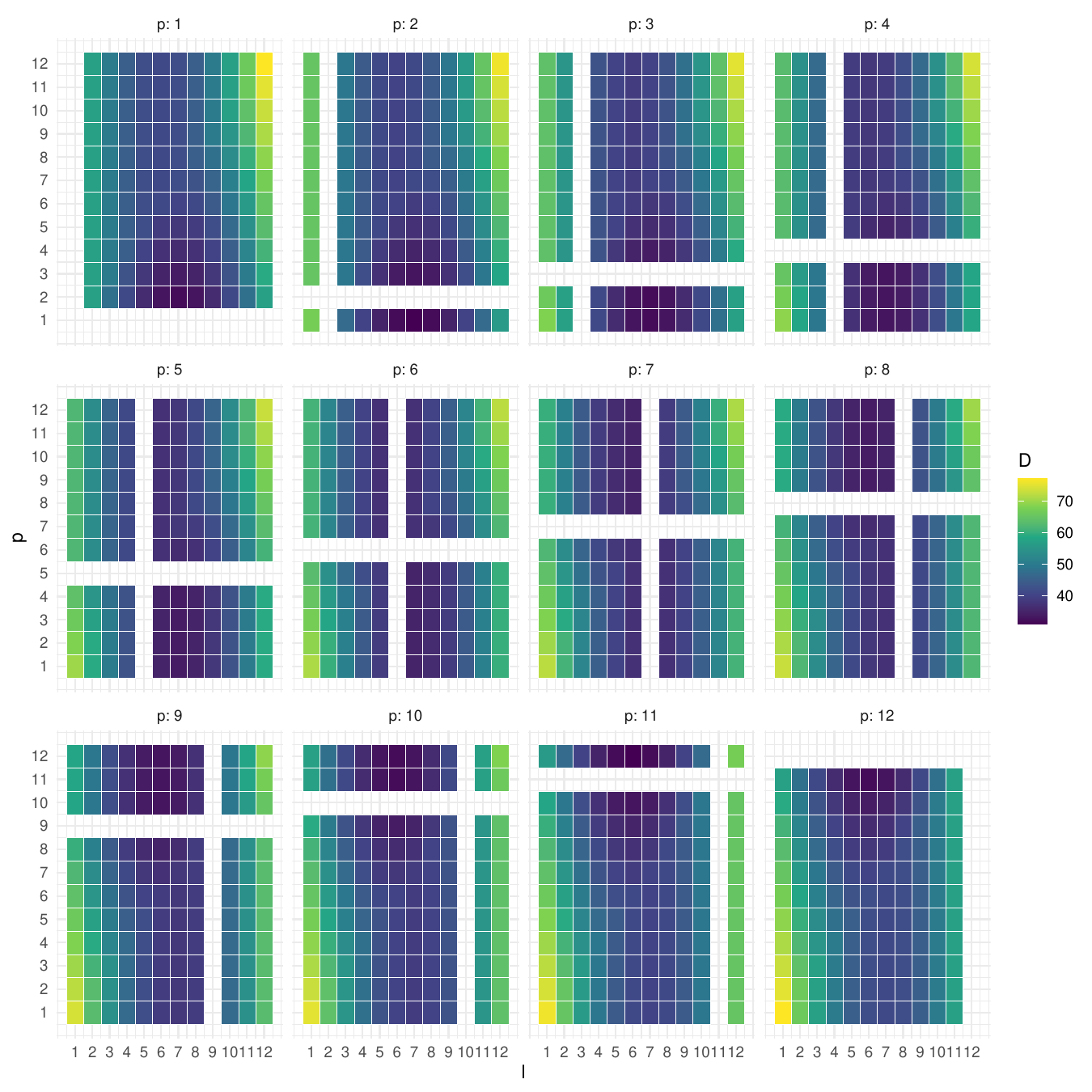}
\caption{\label{fig:landscape_quasistar_by_p} The total sum of dependency distances ($D$) of a quasistar tree with $n = 12$ as a function of $l$, $p$ and $q$. Each panel corresponds to a slice defined by the value of $l$. The format is the same as in Figure \ref{fig:landscape_quasistar_by_q}. }
\end{figure}

The total dependency cost of a quasistar, $D_{qs}(l, p, q)$, is then a three-dimensional function of the positions of the hub ($l$), the node of degree 2 ($p$), and the leaf attached to it ($q$) as shown in Figure \ref{fig:landscape_quasistar_by_q} when slicing by means of $q$ to obtain a two-dimensional heatmap. 
Given just $n$, the range of variation of $D_{qs}$, adapted from \citep[Table 3]{Ferrer2020a}, is
\begin{equation}
\left\lfloor \frac{1}{4} (n - 1)^2 \right\rfloor + 1 \leq D_{qs} \leq \frac{1}{2}(n+3)(n-2)
\label{eq:variation_of_D_quasistar}
\end{equation}
When $n = 12$ as in Figure \ref{fig:landscape_quasistar_by_q}, $31 \leq D_{qs} \leq 75$. See Figure \ref{fig:landscape_quasistar_by_l} for a slicing by means of $l$ and Figure \ref{fig:landscape_quasistar_by_p} for a slicing by means of $p$.

To analyze the convexity of the landscape, we adopt a \textit{rewiring definition}. Let $D_s(n, l)$ be the sum of dependency cosets of a perfect star of $n$ vertices rooted at a hub at position $l$. To form a quasistar tree in a linear arrangement, we disconnect a leaf located at position $q$ from the hub $l$, and reconnect it to a new parent located at position $p$ (where $p$ is one of the other leaves or the hub itself). The total dependency distance cost, $D_q(l, p, q)$, is defined as: 
\begin{equation}
 D_q(l, p, q) = D_s(n, l) - g(|l - q|) + g(|p - q|) \label{eq:rewiring_def} 
\end{equation} 

This definition treats the quasistar as a star with one missing link in the original connection $(l,q)$ and a new link in $(p,q)$. The domain of the variables is $l, q \in [1, n]$ and $p \in [1, n]$.

\subsubsection{Discrete convexity} 
The convexity of $D_q$ is not guaranteed for all cost functions $g(d)$. In Equation \ref{eq:rewiring_def}, we subtract $g(|l-q|)$. If $g(d)$ is convex, $-g(d)$ is concave. The sum of a convex function (the remaining star structure) and a concave term (the removed edge) is not necessarily convex. For the quasistar to remain convex, the convexity of the remaining $n-1$ nodes' contribution must overcome the concavity introduced by the removed edge. More concretely, convexity holds specifically for Linear and Quadratic costs, but is likely to fail for higher-order convex costs (e.g., $g(d) = d^3$ or $e^d$).

\begin{property}
    
For linear cost $g(d)=d$ and quadratic cost $g(d)=d^2$, the quasistar cost function $D_q(l, p, q)$ defines a discretely convex function. \end{property} 

\begin{proof}
     \textbf{Case A: Linear Cost ($g(d) = d$).} The cost function is $D_q = \sum_{i \neq q} |l-i| + |p-q|$. Since the linear function is both convex and concave, subtracting a linear term does not break convexity. 
     
     \textbf{Case B: Quadratic Cost ($g(d) = d^2$).} This works due to an algebraic cancellation of the squared term for the leaf $q$. Expanding the rewiring terms: $$-|l-q|^2 + |p-q|^2 = -(l^2 - 2lq + q^2) + (p^2 - 2pq + q^2)$$ $$= -l^2 + 2lq \mathbf{- q^2} + p^2 - 2pq \mathbf{+ q^2}$$ The $-q^2$ and $+q^2$ terms cancel out. 
     
Therefore the landscape remains convex. 
     \end{proof}
     
     \subsubsection{Analysis of Strong Discrete Convexity} 
     
     While quasistars are discretely convex for linear and quadratic costs, we prove here that they do \textbf{not} satisfy the harder definition of \textit{Strong Discrete Convexity} due to the failure of Local Submodularity. 
     
     However, they satisfy the second condition of strong convexity (Aggregate Monotonicity), which is sufficient to explain the ease of optimization. 
     \begin{property}
      \label{local-submodularity}             
     \textbf{Condition 1: Local Submodularity fails for quasistar trees.} 
              \begin{proof}
Local Submodularity fails when $l$ and $q$ cross each other (so $u=(+1,0,-1$ or $u=(-1, +1, 0)$) in both the Linear and the Quadratic case. 
         \end{proof}
\end{property}

 \begin{property}
     
\textbf{Condition 2: Aggregate Monotonicity holds for quasistar trees.} \begin{proof} Concerning the second condition, we examine the aggregate second differences $\sum_j \Delta_{ij}$. Although the rewiring terms depend on pairs of variables (introducing non-zero mixed differences like $\Delta_{lq}$), their contribution to the aggregate sum vanishes. For example, for quadratic costs, the value added by the term that represents removing an edge is exactly compensated by the positive mixed differences in the summation ($-2$ and $+2$). Consequently, the sign of the aggregate difference is determined solely by the star structure $D_s(n, l)$. Since $D_s(n, l)$ is convex (as proved in Property \ref{prop1}), the aggregate second differences remain non-negative.\end{proof}
\end{property} 
     
 We have proven that quasistar structures satisfy Aggregate Monotonicity (Condition 2 for strongly discrete convexity) for both linear and quadratic costs. While they do not strictly satisfy Local Submodularity (Condition 1), the satisfaction of Aggregate Monotonicity combined with Discrete Convexity is sufficient to support the ease of optimization.

\subsubsection{The planar case} Finally, we consider the specific case where the linear arrangement is constrained to be planar (i.e., no crossing edges). For a quasistar, the subtree rooted at the parent $p$ consists only of $p$ and its dependent leaf $q$. To satisfy planarity, $q$ must be placed immediately adjacent to $p$. This constraint has two decisive effects. First, the distance between $p$ and $q$ is fixed at $|p - q| = 1$. Consequently, the rewiring cost term in Equation \ref{eq:rewiring_def}, $g(|p - q|)$, becomes a constant. The cost function effectively reduces to the cost of a star structure with $n-1$ leaves (treating the $p$-$q$ pair as a single unit). Since star structures define a convex landscape (Section \ref{sec:star_trees}), the landscape of a planar quasistar is also convex. Second, this constraint recovers \textbf{Strong Discrete Convexity}. In the general case, quasistars failed the Local Submodularity condition (Property \ref{local-submodularity}) because the variables $l$ and $q$ could cross each other independently. In the planar case, $q$ loses its independence and moves as a unit with $p$. Since the effective structure is isomorphic to a star, and star trees are strongly discretely convex (Property \ref{strong-convex}), the planar quasistar is also strongly discretely convex.
     
\section{The quadratic cost of star-like structures}
\label{sec:quadratic_cost}

\begin{figure}
\includegraphics[width = \linewidth]{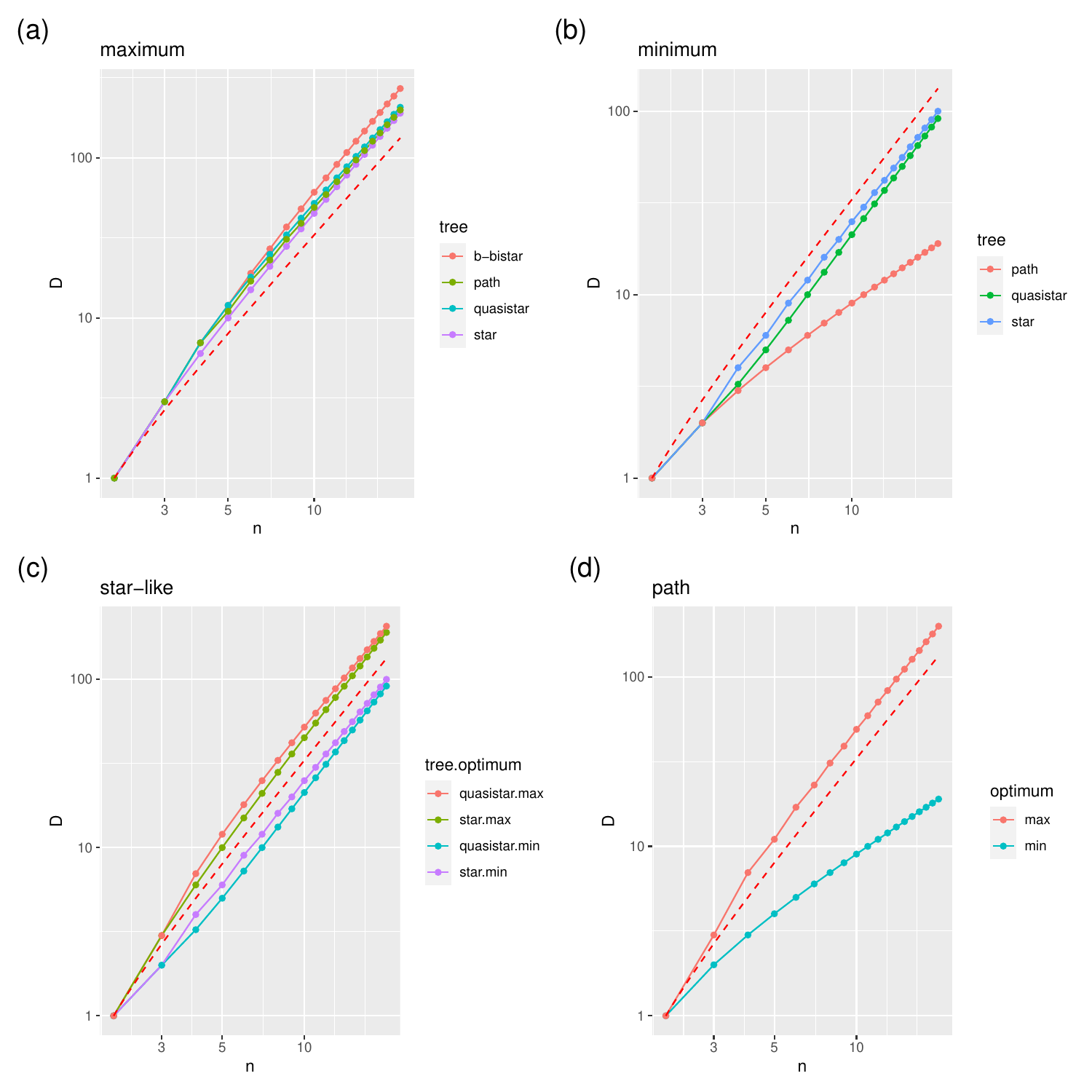}
\caption{\label{fig:quadratic_cost} 
The bounds on the variation the sum of dependency distances, i.e. 
$D$ with $g(d) = d$, as a function of $n$, the size of the tree (sentence length in words). The red dashed line is a control defined by the expected value of $D$ in a uniformly random linear arrangement. $b-bistar$ stands for balanced bistar tree. (a). The lower bound ($D_{min}$) as a function of $n$ for distinct trees. (b). The upper bound ($D_{max}$) as a function of $n$ for distinct trees. (c). The upper and lower bound $D$ ($D_{min}$ and $D_{max}$) in star-like structures (stars and quasistars). (d). The upper and lower bound of $D$ ($D_{min}$ and $D_{max}$) in a path tree. }
\end{figure}

Here we assume $g(d) = d$ to understand the qualitative difference between star-like structures and other structures in terms of dependency distance cost.

\iftoggle{publish}{}
{
\textcolor{blue}{Ramon: notation below to be unified}
}

Consider a dependency structure $t$. We define $D^t$ as its sum of dependency distances in some linear arrangement. We define $D_{min}^t$ and $D_{max}^t$, respectively, as the minimum (min) and the maximum (max) value of $D^t$ over all linear arrangements of $t$. 
We replace $t$ by $s$ for a star tree, $q$ for a quasistar tree and $p$ for a path tree, respectively. 
As a control, we define $D_r^t$, as the expected value of $D^t$ in a uniformly random linear arrangement, that is \citep{Ferrer2018a}
\begin{equation}
D_r^t = \frac{n^2-1}{3}.
\end{equation}
As it only depends on $n$ (not on $t$), we write it simply as $D_r$. 
$D_r$ is a quadratic function of $n$, that is $D_r \sim n^2$.
The range of dependency distance maximization is limited by functions that are also quadratic on $n$ (Figure \ref{fig:quadratic_cost} (a)). In particular, \citep{Ferrer2020a}
\begin{equation}
D_{max, s} = {n \choose 2} \leq D_{max} \leq D_{max, b-bistar} = \frac{1}{4}\left(3(n-1)^2 + 1 - n \bmod 2 \right) \label{eq:variation_of_D_max}
\end{equation}
where $b-bistar$ refers to a balanced bistar tree \footnote{A bistar is a tree formed by joining the hubs of two star trees. A balanced bistar is a bistar where the hubs of the original stars have the same degree of their degree differs by a unit \citep{Ferrer2020a}. }. Therefore, there is no qualitative difference between random order and anti dependency distance maximization.  

The picture changes when considering dependency distance minimization, where the dependence on $n$ is quadratic for star-like-structures but linear for path trees (Figure \ref{fig:quadratic_cost} (b)). The highest gain when minimizing $D$ is obtained by path trees and the lowest gain when minimizing $D$ is achieved by star trees since \citep{Ferrer2020a}
\begin{equation}
D_{min}^p = n - 1 \leq D_{min}^t \leq D_{min}^s = \left\lfloor \frac{1}{4}n^2 \right\rfloor. \label{eq:variation_of_D_min}
\end{equation}
Crucially, there is a qualitative difference between the minima that star trees and path trees achieve. While $D_{min}^p$ is a linear function of $n$, that is $D_{min}^p \sim n$, $D_{min}^s$ (equation \ref{eq:variation_of_D_min}) and $D_{min}^q$ (equation \ref{eq:variation_of_D_quasistar}) are quadratic functions of $n$, i.e. $D_{min}^s,  D_{min}^q \sim n^2$ (Figure \ref{fig:quadratic_cost} (a)). 
Besides, the average value of $D_{min}$ over labelled trees is subquadratic \citep{Esteban2016a} \footnote{\citet{Esteban2016a} considered the average $D_{min}$ of a tree of $n$ vertices, i.e. $\left<d_{min}\right> = D_{min}/(n - 1)$, as a function of $n$ and showed that $\left<d_{min}\right>$ over all unlabelled trees scales as $\log(n)$. Then $D_{min}$ scales as $n \log n$. } 
In his pioneering research, \citet{Iordanskii1974a} derived a subquadratic upper bound of $D_{min}$ for an arbitrary tree with a given maximum vertex degree. 
\footnote{ 
Given a tree of $n$ vertices and maximum degree $k_{max}$, \citet{Iordanskii1974a} found   
\begin{equation}
D_{min} \leq c \frac{k_{max}}{\log k_{max}} n \log n,
\end{equation}
where $c$ is a constant. 
Therefore, we conclude that low gain of star-like structures when minimizing dependency distance, deviates significantly from the majority of structures.
}

To understand the meaning of $D_{min}^s$ and $D_{min}^q$ being quadratic functions of $n$, recall $D_r \sim n^2$ and the always quadratic range of $D_{max}$ in equation \ref{eq:variation_of_D_max}, independently of the tree.
Therefore, a star and a quasistar have minimum and maximum values of $D$ that are quadratic on $n$ (Figure \ref{fig:quadratic_cost} (c)). Therefore, there is no qualitative difference between dependency distance minimization, randomness and dependency distance maximization in star-structures (Figure \ref{fig:quadratic_cost} (c)) while there is indeed such a difference for other structures, e.g. path trees (Figure \ref{fig:quadratic_cost} (d)). 

How the degree star-likeness of a tree determines the gain in dependency distance minimization can be seen easily in caterpillar trees where 
\citep{Ferrer2020a}
\begin{equation*}
D_{min}^t = \frac{1}{4}\left(n\DegreeSecondMoment + z\right),    
\end{equation*}
and $z$ is the number of vertices of odd degree. 
That is, if $z$ remains constant in caterpillar trees, the higher the star-likeness, the higher the value of $D_{min}^t$ and then the lower the gain of dependency distance minimization.

To sum up, it is not surprising that star structures manifest values of dependency distance that are larger than expected by chance \citep{Courtin2019a, Ferrer2019a, Ferrer2023b} as
they are the less rewarding structures in terms of dependency distance minimization. 

\section{Discussion}
\label{sec:discussion} 

\subsection{The research questions}

\subsubsection*{Question b) How difficult is it to minimize the sum of dependency distance in star-like structures?}
 
Here we have shown that the dependency distance minimization landscape of star structures is convex, a more restrictive form of quasiconvexity \citep{Boyd2009a}. Thus, distance-based optimization on star structure is simpler than previously believed \citep{Ferrer2013e}.

we have shown that the landscape remains convex for linear and quadratic costs (Figure \ref{fig:landscape_quasistar_by_q}). An intriguing question is if distance-based optimization on quasistar structures should be simple for that reason. If we take star trees for reference, the number of variables has increased from one (just $l$) to three ($l$, $p$ and $q$), suggesting that the number of variables may increase the complexity of the optimization problem.

\subsubsection*{Question c) Why anti dependency distance minimization effects have been found in star structure but not in path trees?}

A structure with $n = 3$ is both a star tree and a path tree. 
A structure with $n=4$ in either a star tree or a path tree. When $n = 4$, a path tree is equivalent to a quasistar tree. 
Anti-dependency distance minimization effects have been found in structures with $n = 3$ and $n = 4$ \citep{Courtin2019a,Ferrer2019a, Ferrer2023b}.
Could this phenomenon be explained by the complexity of the dependency distance minimization  problem of star trees?
The complexity of the optimization problem depends on the shape of the landscape and the number of parameters to be optimized. When $n = 4$, we have seen that both star trees and path trees, that are also quasistars, have a convex optimization landscape. Besides, the optimization problem in star trees has just one variable ($l$) while the optimization problem in path trees (with $n = 3$) has three variables ($l$, $p$ and $q$) which implies that optimization should be easier in star trees. However, this is contradicted empirically by evidence of dependency distances that are longer than expected by chance in stars but not in paths \citep{Courtin2019a,Ferrer2019a, Ferrer2023b}. Therefore, the reason for the difference between star trees and path trees (with $n=4$) must be structural. In Section \ref{sec:quadratic_cost}, we have shown that there is a critical structural difference in relation to dependency distance minimization. While star-like structures (stars and quasistars) exhibit a quadratic relationship between syntactic dependency distance cost and tree size even when dependency distance is minimum, other structures, e.g. paths, exhibit a maximum cost that is quadratic but a minimum subquadratic cost, which implies a qualitative gain when minimizing syntactic dependency distance. Then, it is not surprising that anti-dependency distance minimization effects appear in star-like structures, where dependency distance minimization is the least rewarding. 

\subsection{Future research}

Here we have shown that the optimization landscape of stars and quasistars is convex. A proof for a wider class of trees should be the subject of future research. Stars and quasistars are particular cases of bistars which is in turn a subset of caterpillar trees.

Here we have relied on empirical research on tree structures with $n=3$, where the trees are stars, and $n=4$, where the trees are either path (or equivalently quasistar with $n=4$) or star \citep{Courtin2019a,Ferrer2019a,Ferrer2023b}. When $n = 5$, the possible trees are star, quasistar and path \citep[Figure 2]{Ferrer2024d}. 
The placement of critical vertices (e.g., the hub) and anti dependency distance effects should be investigated when $n \geq 5$, with special attention to quasistars. An intriguing question for further research is whether, as $n$ increases, there is a bias against star-like structures, not only because of the limited syntactic or semantic expressive power of these structures, but also because their dependency distance minimization costs are high as we have argued here. 


\bibliographystyle{apacite}

\bibliography{bibliography/rferrericancho,bibliography/other}

\end{document}